\documentclass[conference]{IEEEtran}

\IEEEoverridecommandlockouts
\usepackage{cite}
\usepackage{amsmath,amssymb,amsfonts}
\usepackage{algorithmic}
\usepackage{setspace}
\usepackage{graphicx}
\usepackage{textcomp}
\usepackage{xcolor}
\usepackage{float}
\usepackage{url}

\usepackage{xcolor,colortbl}

\definecolor{aliceblue}{rgb}{0.94, 0.97, 1.0}

\usepackage{commath}
\usepackage{ulem}

\usepackage{hyperref}
\hypersetup{
    colorlinks=true,
    linkcolor=blue,
    filecolor=magenta,      
    urlcolor=cyan,
}

\graphicspath{ {.} }

\def\BibTeX{{\rm B\kern-.05em{\sc i\kern-.025em b}\kern-.08em
    T\kern-.1667em\lower.7ex\hbox{E}\kern-.125emX}}
\begin{document}

\title{ A Study On the Effects of Pre-processing On Spatio-temporal Action Recognition Using Spiking Neural Networks Trained with STDP\\
}
\author{
\IEEEauthorblockN{
Mireille El-Assal\IEEEauthorrefmark{1},
Pierre Tirilly\IEEEauthorrefmark{1},
and Ioan Marius Bilasco\IEEEauthorrefmark{1}
}
\IEEEauthorblockA{\IEEEauthorrefmark{1}
\textit{Univ. Lille, CNRS, Centrale Lille,}
\textit{UMR 9189 -- CRIStAL -- Centre de Recherche en Informatique, Signal et Automatique de Lille}\\
F-59000, Lille, France\\
}
\IEEEauthorblockA{
Email: mireille.elassal2@univ-lille.fr, pierre.tirilly@univ-lille.fr, marius.bilasco@univ-lille.fr
}
}
\maketitle
\begin{abstract}
There has been an increasing interest in spiking neural networks in recent years. SNNs are seen as hypothetical solutions for the bottlenecks of ANNs in pattern recognition, such as energy efficiency \cite{b1}. But current methods such as ANN-to-SNN conversion and back-propagation do not take full advantage of these networks, and unsupervised methods have not yet reached a success comparable to advanced artificial neural networks. It is important to study the behavior of SNNs trained with unsupervised learning methods such as spike-timing dependent plasticity (STDP) on video classification tasks, including mechanisms to model motion information using spikes, as this information is critical for video understanding. This paper presents multiple methods of transposing temporal information into a static format, and then transforming the visual information into spikes using latency coding. These methods are paired with two types of temporal fusion known as early and late fusion, and are used to help the spiking neural network in capturing the spatio-temporal features from videos. In this paper, we rely on the network architecture of a convolutional spiking neural network trained with STDP, and we test the performance of this network when challenged with action recognition tasks. Understanding how a spiking neural network responds to different methods of movement extraction and representation can help reduce the performance gap between SNNs and ANNs. In this paper we show the effect of the similarity in the shape and speed of certain actions on action recognition with spiking neural networks, we also highlight the effectiveness of some methods compared to others.
\end{abstract}


\section{Introduction}
Spiking neural networks are biologically-inspired third generation neural networks modelled after the human brain \cite{b8}. In these networks the communication between neurons is done by broadcasting spike trains. Some of their advantages over ANNs are biological plausibility, fast information processing when implemented on dedicated hardware, and energy efficiency \cite{b8}, \cite{ b10, b20}, not to mention that spiking events are sparse in time, which means that these spikes can potentially hold a large amount of information \cite{b8}. Despite all of these advantages, and the many theories in which SNNs are capable of avoiding certain bottlenecks of ANNs, current methods such as spatio-temporal back-propagation \cite{b17} and ANN-to-SNN conversion \cite{b21} do not completely overcome the bottlenecks of ANNs nor certain SNN related limits, such as frequency loss \cite{b5}. On the other hand, models trained with spike-timing dependent plasticity (STDP) allow local computations, thus enabling the implementation of these networks on larger ranges of devices, but these models still do not compete with the results achieved by ANNs \cite{b3}. Human action recognition is a standard computer vision problem, that can be addressed with many neural network models. Because of its wide range of applications, it is valuable to challenge a spiking neural network with video analysis tasks, in order to inspect the ability of SNNs in processing visual information. But spiking models are still far behind traditional models. Therefore, in order for SNNs to stand out, there is a need for unsupervised methodologies that can make them effectively learn spatio-temporal features, because unsupervised learning has the ability to develop new biologically plausible self-learning methods without needing excessive amounts of labeled data. 

In this work, we aim to learn spatio-temporal features in an unsupervised manner with STDP, by pairing different pre-processing methods with two temporal fusion methods, thus, generating static representations that encode local movement. After that, we evaluate the performance achieved by these representations on a convolutional SNN. Experiments are performed on the KTH and Weizmann datasets, which are natural datasets similar to real live applications in computer vision; although ideal recognition rates have already been achieved on these datasets using traditional computer vision approaches, their simplicity makes them good basic benchmarks to study the performance of new models like SNNs trained with STDP when challenged with spatio-temporal information. Thus, attempting unsupervised feature learning using STDP on these datasets is a first step towards bridging the performance gap between SNNs and other deep learning solutions.

\section{Related Work}
\noindent \textbf{Common spiking neural network learning methods.} The most common spiking neural network models featured in the literature are based on ANN-to-SNN transformation, and supervised learning techniques, such as adapting back-propagation on SNNs \cite{b4, b13}, which target high recognition rates, setting aside many of the advantages of SNNs, such as energy efficiency during training on dedicated hardware. Some of these models are described in this section. An ANN-to-SNN transformation is applied in \cite{b19} where a regular ANN is trained for a given sequence of input frames, and streaming rollouts are used to compute the activations of all ANN units over time. They applied back-propagation-through-time in order to train their network, and then they transformed their ANN into an SNN. However, in their work they do not address the ANN bottlenecks SNNs were made to avoid, because they use a regular ANN to conduct the training. A supervised approach is also suggested in \cite{b17}, where the authors use a supervised spatio-temporal back-propagation (STBP) algorithm for training SNNs. They solved the "non-differentiable" problem caused by the nature of spikes by using surrogate gradients as approximate derivatives for spike activity. But, in their approach, each convolutional neuron receives pre-prepared convoluted results as input, thus using heavy and costly pre-processing instead of training the convolutional kernels from scratch with spike information. SPAN \cite{b2} is a spiking neural network used to classify spatio-temporal data by transforming spike trains during the learning phase into analog signals. In \cite{b2}, the authors are able to successfully teach their system to recognise certain simple patterns of numbers that they created. But they did not conduct any experiments on a reference video dataset. Therefore, there is no evidence that their model would be realistically applicable on more complex datasets. Another SNN learning method is the BCM (Bienenstock, Cooper, and Munro) learning rule. In \cite{b22}, the authors proposed a BCM-based spiking neural network model that classifies human action recognition videos. Another learning rule is STDP, which is a biologically plausible unsupervised learning rule \cite{b6}. In \cite{b23}, the authors use STDP learning on a deep SNN. They use temporal coding, and train their network on natural images for the sake of object recognition. In \cite{b24} the authors use reward-modulated Spike-Timing-Dependent-Plasticity (R-STDP) and reinforcement learning to train their network to perform action classification. In \cite{b25}, the authors use a supervised reward-modulated Spike-Timing-Dependent-Plasticity (R-STDP) learning rule to train two SNN-based sub-controllers on obstacle avoidance tasks. In this work we explore another way of training the network to perform action classification. We use the biological STDP learning rule \cite{b15} in an unsupervised manner to train our convolutional spiking neural network to learn spatio-temporal features. This unsupervised learning gives the advantage of not needing a large amount of labeled data.

\noindent \textbf{Spatio-temporal information learning.} In this section, we review traditional models, such as ConvNets, that can learn spatio-temporal information from videos. In \cite{b27}, the authors evaluate multiple approaches of extending CNNs into video classification. Then they highlight an architecture that separates the spatial information of the input into a low-resolution and a high-resolution context stream. After that, they describe multiple fusion methods to fuse the information across the temporal domain. In \cite{b16}, a two-stream model is introduced. In this model, two deep convolutional networks are used to separate the spatial and temporal recognition streams. The spatial stream relies on still frames and is responsible for the information regarding appearance, while the temporal stream relies on multi-frame dense optical flow and is responsible for the movement information found in the motion between frames. These two streams are combined by a type of feature fusion which is late fusion \cite{b9}. This fusion forms the complementary information needed to achieve action recognition in videos. A more complex approach is explained in \cite{b7}, where a spatio-temporal pyramid architecture is introduced. In this paper, the authors used an architecture similar to the two-stream method in \cite{b16} in the first stage. The spatial stream is represented by still RGB frames that contain the appearance information, while optical flow is used to capture the motion between frames, in the temporal stream. Then these channels are fused together in the first step, as in two-stream methods, but a multi-level fusion pyramid of spatio-temporal features is added when the same streams are fused again in step two with the result of their previous fusion. Although not implemented in the context of SNNs, these models are interesting because they show the importance of both the spatial as well as the temporal information in action recognition. They also highlight the importance of feature fusion in creating a complete data representation. Temporal fusion is introduced in \cite{b9} where the authors create a Dual Temporal Scale Convolutional Neural Network (DTSCNN) architecture to recognize spontaneous micro-expression.

In brief, there is a need for understanding different methods that represent the spatio-temporal information found in videos, and their implementation with SNNs. This work can serve as a first step towards creating models that can learn spatio-temporal features and conduct their training locally, in a processing-cost friendly manner that can be used in real-world applications. This energy efficiency can be achieved with STDP learning. This paper contributes to the study of bringing closer the nature of the spatio-temporal information and the nature of STDP trained spiking neural network, in order to insure better performance.

\section{Network Architecture}

\begin{figure}
\centerline{\includegraphics[scale=0.3]{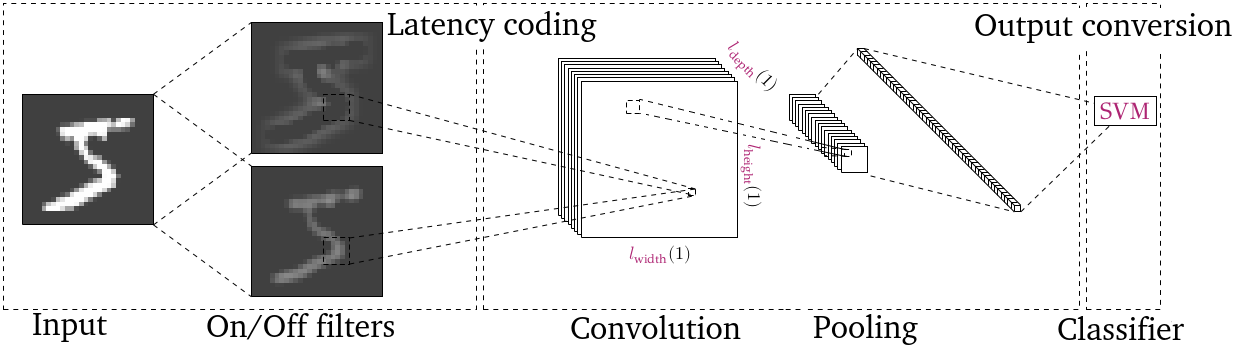}}
\caption{Network topology (figure from \cite{b15}).}
\label{fig1}
\end{figure}

\noindent \textbf{The general architecture.}
In this paper we use a state-of-the-art convolutional SNN model from \cite{b15} which consists of feed-forward layers that contain IF neurons \cite{b14}, and trained using the biological STDP learning rule \cite{stdprule}. An on-center/off-center filter is used to pre-process the data before latency coding is applied to transform this data into spikes. The threshold adaptation method described in \cite{b15} is used in order to maintain a state of homeostasis. The SNN we chose uses only one (convolution/pooling) stage for simplicity, as shown in (Fig. \ref{fig1}). This is because training multi-layer SNNs with STDP is still an open problem \cite{b10}. The objective of this paper is to focus on how spatio-temporal data can be pre-processed in order to feed a convolutional SNN with temporal information aggregated from a video sequence. We also explore early and late fusion techniques \cite{b9} in order to evaluate the benefits of such techniques in encoding spatio-temporal information. The output of this network is a dense array that represents the processed sample, flattened as a linear array and introduced into a support vector machine (SVM) that is used to make the action classification. An SVM is used to classify the samples because we focus on the unsupervised learning of features. Any other supervised method can be used to do the classification, but we chose to use an SVM for its simplicity and its effectiveness.

\begin{figure}[H]
\centerline{\includegraphics[scale=0.35]{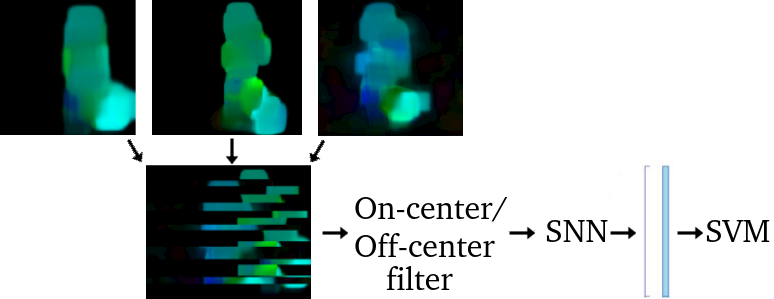}}

\caption{Early fusion, multiple input frames are fused together.}
\label{fig2}
\end{figure}

\begin{figure}
\centerline{\includegraphics[scale=0.3]{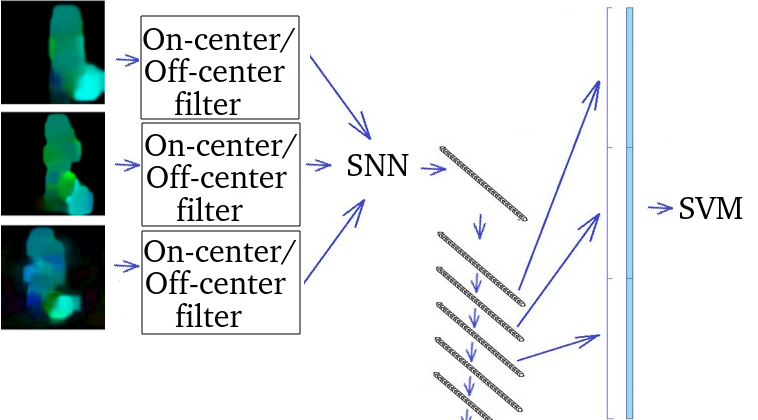}}
\caption{Late fusion: the features obtained at the end of the last pooling layer are fused together before entering the SVM.}
\label{fig3}
\end{figure}

\noindent \textbf{Feature fusion.}
Temporal fusion is one way of aggregating the temporal information in a sequence of frames. In deep neural networks, there are multiple types of fusion \cite{b9}, but in this paper, only early and late fusions are studied and implemented. They are a good fit to our study because we are using a single layer SNN model, which makes implementing other fusion methods that require multiple layers inapplicable, such as slow fusion \cite{b9}. Early and late fusion techniques operate differently and are implemented separately in this work. Early fusion is implemented by taking multiple samples and fusing them together row by row, into one big frame, as shown in the following equation: $I^o_{kj} = I^f_{ij}$ with $k=i*n+f$, $ f \in [0,n-1], i \in [0,h-1], j \in [0,w-1]$, where $I^f$ is the input frame of index f, and $I^o$ is the output frame. On the other hand, late fusion is implemented by taking multiple flattened samples at the output of the network and concatenating them together.
The main difference between early and late fusion is the stage at which the fusion takes place. In early fusion, shown in (Fig. \ref{fig2}), the sample frames are fused together before training the convolutional kernels. On the other hand, in late fusion, see (Fig. \ref{fig3}), the features that result from the processing of the video frames are fused together in the last stage. In the late fusion method implemented in this work, the samples are flattened and then joined together using a sequential queue.

\begin{figure}[H]
\centerline{\includegraphics[scale=0.32]{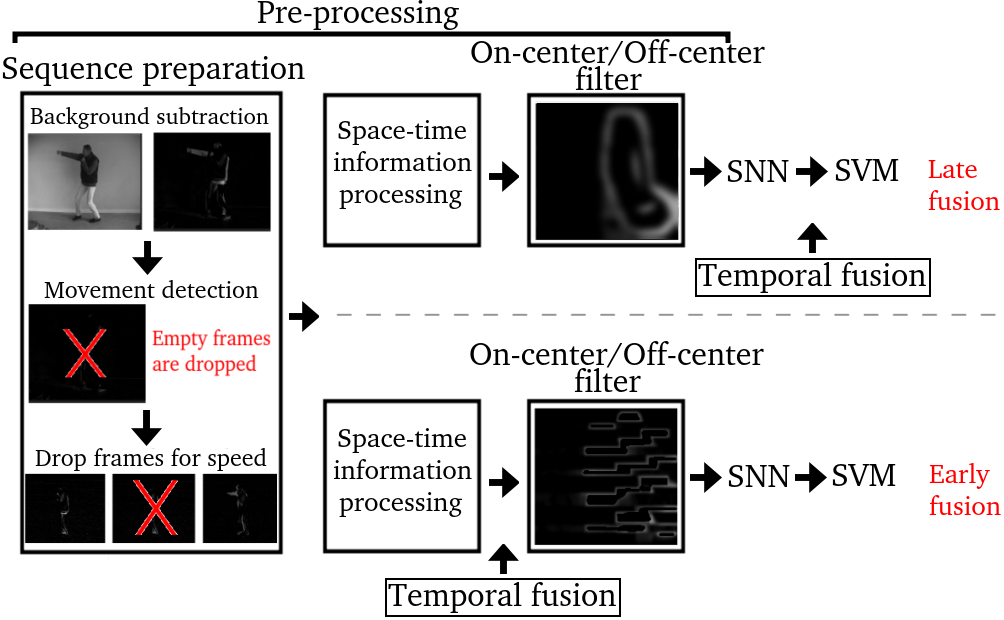}}
\caption{The general pre-processing procedure.}
\label{fig4}
\end{figure}

\noindent \textbf{The process.}
The general sequence preparation (SP) is a pre-processing procedure (see Fig. \ref{fig4}) that consists in applying background subtraction to every two consecutive frames, because it reduces the noise that can results from optical flow computation. After that, frames that do not contain significant motion are dropped. This is done by averaging the values of pixels and checking if this average is greater than a threshold estimated by trial and error, where each method has its own threshold. Two frames are also dropped between every two frames that are selected to be used in the sequence; this speeds up the action and helps recording it in a relatively smaller number of frames. Then, a pre-processing method based on Farneback's dense optical flow \cite{b18} is applied. Early fusion can be applied directly after implementing the optical flow representation, unless late fusion is going to be applied at the end of the process. Then, on-center/off-center filtering is applied and the data is introduced into the convolutional SNN.

\section{Space-Time Information Processing}
In this section, we present different pre-processing methods for action recognition video datasets, and evaluate their suitability to spatio-temporal feature learning with STDP-based SNNs. We rely on deriving different data representations from sequences of static frames. We base our pre-processing on Farneback's dense optical flow because it is a reference method in motion representation. We define five representations that exploit various aspects of the optical flow: the horizontal and vertical displacement (DXDY), the Orientation and Amplitude (OA), the Composite Channel information (CC), the Edges Grid (EG) and the Motion Grid (MG). The initial processes described in Section III-C are used for all the pre-processing methods except the EG and MG methods. 

\begin{figure}
\centerline{\includegraphics[scale=0.3]{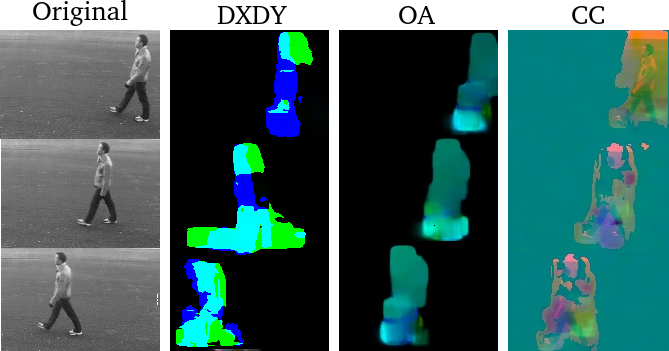}}
\caption{A Walking action. (A) The original frame. (B) The DXDY representation (in RGB, G: $D_x$, B: $D_y$). (C) The OA representation(in HSV, H: orientation, V:amplitude). (D) The CC representation (in RGB, R: the moving part of the original grey-scale image G: $D_x$, B: $D_y$).}
\label{fig5}
\end{figure}

\noindent \textbf{The DXDY representation.} DXDY is made up of 2 channels, a horizontal displacement $D_{x}$ in channel $C_{1}$ and a vertical displacement $D_{y}$ in channel $C_{2}$. A sample produced by using this method is shown in (Fig. \ref{fig5}(B)).

\noindent \textbf{The OA representation.} OA separates optical flow vectors into orientation and magnitude values. The orientation data is periodic, therefore, it is difficult to apply latency coding to it. Thus, the information is displayed in the HSV color space, which is then converted into RGB color space (see Fig. \ref{fig5}(C)).

\noindent \textbf{The CC representation.} CC is created by combining the channels of the DXDY representation with the gray scale appearance information of the moving subject. A sample that results from this method contains three channels (see Fig. \ref{fig5}(D)). The first channel $C_{1}$ represents the horizontal displacement $D_{x}$, the second channel $C_{2}$ represents the vertical displacement $D_{y}$, and the third channel $C_{3}$ represents the original gray scale information of only the moving parts of the subject. The gray scale illumination of each pixel corresponds to the mean value of the channels in the original image, and pixels with significant motion are detected as follows: $\abs{D_{x}} + \abs{D_{y}} > \theta $, we use $\theta = 30$ in the experiments.

\begin{figure}[H]
\centerline{\includegraphics[scale=0.55]{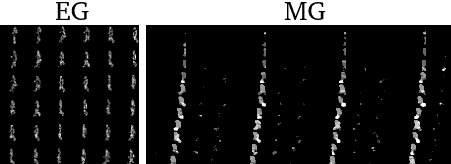}}
\caption{A Walking action. (A) EG representation. (B) MG representation.}
\label{fig6}
\end{figure}

\noindent \textbf{The EG representation.} EG is based on extracting the edges of motion from optical flow frames using the Canny edge detection approach \cite{b28}. In spirit, the EG representation resembles motion boundary descriptors \cite{b12}, except that we group these  edges of motion into a grid as shown in Fig. \ref{fig6}(A). Each sample is constructed using 36 optical flow frames, and sample frame overlapping is used to increase the number of samples. Many different grid sizes were tested but 36 proved to be the most suitable value. This creates an early fusion of 36 frames, and therefore, the feature fusion methods in Section III-B are not applied with this method.

\noindent \textbf{The MG representation.} MG groups the movement information into a composite grid that is made up of $4\times12$ optical flow frames. Each frame is divided into 4 channels that are placed in separate frames one after the other as shown in (Fig. \ref{fig6}(B)). These channels are: the horizontal displacement to the left $\abs{- D_{x}}$, the horizontal displacement to the right $+ D_{x}$, the vertical displacement in the upwards direction  $\abs{- D_{y}}$, and the vertical displacement in the downwards direction $+ D_{y}$, resulting in $16\times12$ channels per grid. This creates an early fusion of 48 frames, therefore, the feature fusion methods in Section III-B are also not applied with this method.

\section{Evaluation}
\noindent \textbf{Data Sets.}
The \href{https://www.csc.kth.se/cvap/actions/}{KTH} dataset \cite{kth1} contains 600 videos of 25 subjects, performing 6 actions in 4 scenarios. The subjects 11, 12, 13, 14, 15, 16, 17 and 18 are used for training, while 02, 03, 05, 06, 07, 08, 09, 10 and 22 are used for testing, as indicated in the KTH protocol. The \href{http://www.wisdom.weizmann.ac.il/~vision/SpaceTimeActions.html}{Weizmann} dataset \cite{weizmann1} contains 90 videos of 9 subjects performing 10 actions. The experiments on this dataset are done using the leave-one-out strategy. Each sample from both datasets has 10 frames prepared as described in Section III-C. This applies for all representations, except the MG and EG representations which need 48 and 36 frames per sample respectively.

\noindent \textbf{Meta-parameters of the Model.}
The meta-parameters used in this work are presented in Table \ref{table:I}. A difference-of-Gaussian (DoG) filter is used to simulate on-center/off-center cells. This filter creates a motion boundary effect \cite{b12} that increases the classification rates. Experiments with and without this filter were conducted, and experiments without this filter gave inferior results. Different numbers and sizes of filters were tested, but we only reported the most suitable values: 128 convolutional kernels of size 5 x 5, with a padding of 2 and a stride of 1. The convolutional SNN tested in this work is simulated using the \href{https://gitlab.univ-lille.fr/bioinsp/falez-csnn-simulator/tree/07fd14324afc42d7b3b24a3472271e1c6a90255a}{falez-csnn-simulator} \cite{b15}. 

\begin{table}[H]
\begin{center}
 \begin{tabular}{| c |} 
 \hline
 \rowcolor{aliceblue}\textbf{Learning} \\
 \hline
 $\alpha= 0.95$,  $n_{epoch}= 100$ \\
 \hline
  \rowcolor{aliceblue}\textbf{STDP} \\
 \hline
 $W_{min}= 0.0$,  $W_{max}= 1.0$, $\eta_{w(0)}= 0.1$, \\ $	\beta= 1.0$, $\tau_{STDP}= 0.1$, $w(0) \sim U(0, 1)$ \\
 \hline
  \rowcolor{aliceblue}\textbf{Neural Coding} \\
 \hline
 $\mathrm{t_{exposition}}=$ 1.0 \\
 \hline
 \rowcolor{aliceblue}\textbf{Threshold Adaptation }\\
 \hline
 $\mathrm{t_{expected}}= 0.95$, $\eta_{\theta}(0)= 1.0$, $th_{min}= 1.0 $, \\ $\upsilon_{\theta}(0) \sim G(5, 1)$, $\upsilon_{inh}= 1.0$ \\
 \hline
 \rowcolor{aliceblue}\textbf{Difference-of-Gaussian} \\
 \hline
 $\mathrm{DoG_{center}}= 1.0$, $\mathrm{DoG_{surround}}= 4.0$, $\mathrm{DoG_{size}}= 7.0$ \\
 \hline
\end{tabular}
\end{center}
\caption{The meta-parameter values used in the expirements. See \cite{b15} for notations.}
\label{table:I}
\end{table}

\noindent \textbf{Baseline. }
We cannot compare our evaluation with the state-of-the-arts because, to the best of our knowledge, none of the previous work in the literature use STDP with unsupervised learning for video classification as mentioned in Section II. Table \ref{table:II} displays the classification rates obtained by training the convolutional SNN using raw video frames. Each sequence fused using early and late fusion is made up of $10$ frames. The sequence preparation (SP) process mentioned in section (III-C) is applied. This table serves as a baseline in order to compare the effectiveness of the pre-processing techniques.

\begin{table}[H]
\begin{center}
 \begin{tabular}{|c | c | c | c | c |} 
 \hline
 \rowcolor{aliceblue}Dataset & KTH & Weizmann & KTH + SP & Weizmann + SP \\ [0.5ex] 
 \hline
 No Fusion & 19.54 & 18.88 & 34.84 & \textbf{24.49}\\ [0.25ex] 
 \hline
 Early Fusion  & 24.50 & 22.11 & 30.52 & 20.73 \\ [0.25ex] 
 \hline
 Late Fusion &26.38 & 21.03 &  \textbf{35.26} & 23.58 \\ [0.25ex] 
 \hline
\end{tabular}
\end{center}
\caption{ Classification rate in \% using early, late, and no fusion with the KTH and Weizmann datasets as raw frames, with and without sequence preparation (SP).}
\label{table:II}
\end{table}

Sequence preparation yields higher classification rates. This is because background subtraction removes some unnecessary spatial information, such as clothing and surrounding objects (see Fig.~\ref{fig4}). Without sequence preparation, temporal fusion increases the classification rate with respect to no fusion. This is due to the SNN depending on the temporal information to classify the action. But early fusion decreases the classification rate after sequence preparation. This is because the SNN is no longer confused by the extra spatial information, and only considers the form of the action. Another reason is that fusion decreases the number of training samples by $10$, since every $10$ frames are processed as one sample.

\noindent \textbf{Evaluation of the Five Representations.}
Preparing pre-processed samples from the action recognition videos using the DXDY, OA, CC and MG representations, yields the results displayed in Table \ref{table:III}.

\begin{table}[H]
\begin{center}
\begin{tabular}{|c | c | c | c | c|} 
\hline
\rowcolor{aliceblue}Dataset  & \multicolumn{2}{c|}{KTH} & \multicolumn{2}{c|}{Weizmann} \\
\hline
Fusion Method & Early & Late & Early & Late \\ [0.5ex] 
\hline
DXDY  & 26.85 & 28.70  & 12.86 & 11.11  \\ [0.25ex] 
\hline
Orientation and Amplitude  & 41.05 & 45.83  & \textbf{50.42} & 44.44  \\ [0.25ex] 
\hline
Composite Channels  & 45.78 & 54.21  & 36.75 & 30.68 \\ [0.25ex] 
\hline
Edges Grid  & 63.01 & -  & 43.83 &- \\ [0.25ex] 
\hline
Motion Grid  & \textbf{77.69} & -  &28.86 &- \\ [0.25ex]
\hline

\end{tabular}
\end{center}
\caption{Classification rate in \% using early fusion and late fusion with the KTH and Weizmann datasets as pre-processed frames.}
\label{table:III}
\end{table}

\begin{figure}
\centerline{\includegraphics[scale=0.29]{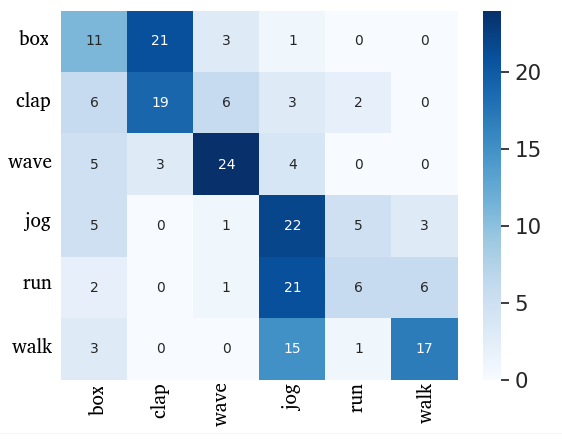}\includegraphics[scale=0.29]{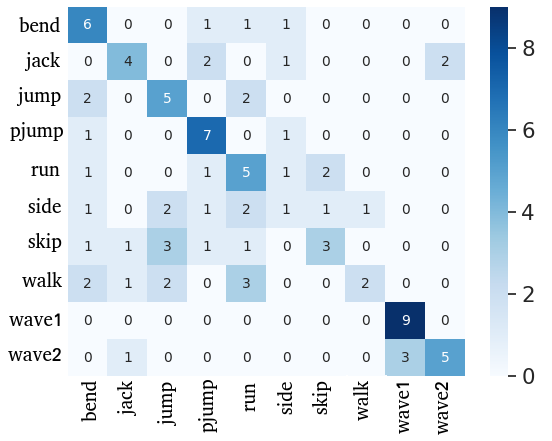}}
\caption{Confusion Matrix of the (A) KTH dataset and (B) Weizmann dataset pre-processed with the OA method and introduced into the SNN.}
\label{fig7}
\end{figure}

DXDY depends only on the displacement information. The datasets contain some actions that are similar in form, and others that are similar in their amount of displacement. Thus, DXDY is not efficient enough to discriminate actions. The OA representation gives a slightly higher classification rate. With this method, the SNN is able to learn that the features of the first set of three actions Boxing, Clapping, and Waving are different from the other set of three actions Jogging, Walking, and Running (see Fig. \ref{fig7}), but actions that are relatively similar in form are not well discriminated. The biggest confusion was recorded between the Jogging and Running actions, which are similar. The same applies to the Weizmann dataset, where the actions Bend and Jack are well differentiated, while there is confusion between similar actions such as Walk and Run. The CC representation aims to add spatial information to the movement information, and shows a slightly better performance than the OA method in the case of KTH dataset. 

The MG outperforms all the other pre-processing methods when implemented with the KTH dataset (see Table \ref{table:III}). This is because it densely represents the temporal movement information. On the other hand, the performance of this method on the Weizmann dataset is poor, which is due to the lack of training samples. The Weizmann dataset videos are scarcer and have shorter durations than the KTH ones, and each MG sample requires 48 frames, thus not enough training samples can be generated from the Weizmann dataset using this method. To highlight this, we increased the number of training samples with the MG representation from 110 to 220 by horizontally flipping the videos. Then we added Gaussian noise to the set of frames, thus doubling the number of samples again, to 440 training samples. The classification rate increased to 45\%. We also experimented with decreasing numbers of samples on the KTH dataset, and the accuracy decreased accordingly. Finally, the EG representation gives an inferior classification rate to the MG representation when using the KTH dataset. On the other hand, using the Weizmann dataset with the EG representation gives a higher classification rate than the MG representation. This is because fusing 36 frames per sample along with sample overlapping generates more samples than fusing 48 frames per sample: here again, the number of training samples in critical in reaching good performances. It is important to note that we also tested the MG method applied on the KTH dataset with a regular CNN. This test gave a classification rate of (77\%), which is similar to that obtained with our SNN. Although the MG does not give results that are state-of-the-art in comparison to the classification rates obtained by other methods in testing human action recognition datasets with ANNs, it does show potential in understanding how SNNs may handle spatio-temporal information. 

\section{Discussion}
Using multiple pre-processing techniques in addition to early and late fusion methods helps in understanding how SNNs can achieve human action recognition. The velocity distribution of the moving components in the videos and the shapes of these components are two very important aspects in action classification. When using a pre-processing method that clearly highlights these two aspects, the spiking neural network was able to reach higher classification rates. The MG grid is able to represent at least one complete cycle of the action being performed with the KTH dataset, thus forming a more complete data representation. Spiking neural networks are still not able to achieve the recognition rates that regular convolutional neural networks can achieve with action classification tasks, and therefore more research needs to be conducted in this field. As a result it may be a good idea to implement the two-stream method \cite{b16} which is similar in spirit to the CC representation, except that in this method, spatial information has an entire dedicate stream, while in the CC method, the spatial information is processed by the same neurons. Another idea would be to implement the same experiments with a multi-layer spiking neural network, which is a very challenging task \cite{b10}. 

\section{Conclusion}
This study was carried out in order to give an assessment of the effect of different data representations on spatio-temporal feature learning. The result of testing these representations on an SNNs trained with STDP yields several conclusions. The first conclusion is that the spatial information improves the classification rate with respect to using the displacement information alone. The second conclusion is that the best action classification rate is recorded when there is at least a full cycle of motion, like in the case of the MG representation with KTH dataset videos. The same MG representation gave inferior results using the Weizmann dataset, because the videos are not long enough to fill this grid with multiple cycles of motion (some videos contain only one action, others contain two, etc.), or to create enough samples. Testing the MG representation with regular 2D CNNs gives a similar recognition rate, which proves that a suitable pre-processing method can help bridge the gap between SNNs and CNNs. The MG shows an improvement compared to the other methods of pre-processing, and serves as a good starting point in improving human action recognition with SNNs.

\section*{Acknowledgments}
This work has been partially funded by IRCICA (USR 3380) under the bio-inspired project.

\bibliographystyle{IEEEtran} 
\bibliography{preprocessing}

\begin{thebibliography}{10}
\providecommand{\url}[1]{#1}
\csname url@samestyle\endcsname
\providecommand{\newblock}{\relax}
\providecommand{\bibinfo}[2]{#2}
\providecommand{\BIBentrySTDinterwordspacing}{\spaceskip=0pt\relax}
\providecommand{\BIBentryALTinterwordstretchfactor}{4}
\providecommand{\BIBentryALTinterwordspacing}{\spaceskip=\fontdimen2\font plus
\BIBentryALTinterwordstretchfactor\fontdimen3\font minus
  \fontdimen4\font\relax}
\providecommand{\BIBforeignlanguage}[2]{{%
\expandafter\ifx\csname l@#1\endcsname\relax
\typeout{** WARNING: IEEEtran.bst: No hyphenation pattern has been}%
\typeout{** loaded for the language `#1'. Using the pattern for}%
\typeout{** the default language instead.}%
\else
\language=\csname l@#1\endcsname
\fi
#2}}
\providecommand{\BIBdecl}{\relax}
\BIBdecl

\bibitem{b1}
S.~Ghosh-Dastidar and H.~Adeli, ``Third generation neural networks: Spiking
  neural networks,'' in \emph{Advances in Computational Intelligence}, W.~Yu
  and E.~N. Sanchez, Eds.\hskip 1em plus 0.5em minus 0.4em\relax Berlin,
  Heidelberg: Springer Berlin Heidelberg, 2009, pp. 167--178.

\bibitem{b8}
A.~Tavanaei, M.~Ghodrati, S.~R. Kheradpisheh, T.~Masquelier, and A.~Maida,
  ``Deep learning in spiking neural networks,'' \emph{Neural Networks}, vol.
  111, pp. 47 -- 63, 2019.

\bibitem{b10}
P.~{Falez}, P.~{Tirilly}, I.~{Marius Bilasco}, P.~{Devienne}, and P.~{Boulet},
  ``Multi-layered spiking neural network with target timestamp threshold
  adaptation and stdp,'' in \emph{2019 International Joint Conference on Neural
  Networks (IJCNN)}, 2019, pp. 1--8.

\bibitem{b20}
Z.~Bing, I.~Baumann, Z.~Jiang, K.~Huang, C.~Cai, and A.~Knoll,
  ``\BIBforeignlanguage{English}{Supervised {Learning} in {SNN} via
  {Reward}-{Modulated} {Spike}-{Timing}-{Dependent} {Plasticity} for a {Target}
  {Reaching} {Vehicle}},'' \emph{\BIBforeignlanguage{English}{Frontiers in
  Neurorobotics}}, vol.~13, 2019.

\bibitem{b17}
Y.~Wu, L.~Deng, G.~Li, J.~Zhu, and L.~Shi, ``Spatio-temporal backpropagation
  for training high-performance spiking neural networks,'' \emph{Frontiers in
  Neuroscience}, vol.~12, 2018.

\bibitem{b21}
B.~Rueckauer, I.~Lungu, Y.~Hu, M.~Pfeiffer, and S.-C. Liu, ``Conversion of
  continuous-valued deep networks to efficient event-driven networks for image
  classification,'' \emph{Frontiers in Neuroscience}, vol.~11, 2017.

\bibitem{b5}
P.~{Falez}, P.~{Tirilly}, I.~M. {Bilasco}, P.~{Devienne}, and P.~{Boulet},
  ``Mastering the output frequency in spiking neural networks,'' in \emph{2018
  International Joint Conference on Neural Networks (IJCNN)}, 2018, pp. 1--8.

\bibitem{b3}
P.~Falez, P.~Tirilly, I.~M. Bilasco, P.~Devienne, and P.~Boulet, ``Unsupervised
  visual feature learning with spike-timing-dependent plasticity: How far are
  we from traditional feature learning approaches?'' \emph{Pattern
  Recognition}, vol.~93, p. 418–429, 2019.

\bibitem{b4}
A.~V. {Gavrilov} and K.~O. {Panchenko}, ``Methods of learning for spiking
  neural networks. a survey,'' in \emph{2016 13th International
  Scientific-Technical Conference on Actual Problems of Electronics Instrument
  Engineering (APEIE)}, vol.~02, 2016, pp. 455--460.

\bibitem{b13}
C.~Lee, S.~S. Sarwar, P.~Panda, G.~Srinivasan, and K.~Roy, ``Enabling
  spike-based backpropagation for training deep neural network architectures,''
  \emph{Frontiers in Neuroscience}, vol.~14, 2020.

\bibitem{b19}
A.~Kugele, T.~Pfeil, M.~Pfeiffer, and E.~Chicca, ``Efficient {Processing} of
  {Spatio}-{Temporal} {Data} {Streams} {With} {Spiking} {Neural} {Networks},''
  \emph{Frontiers in Neuroscience}, vol.~14, 2020.

\bibitem{b2}
A.~Mohemmed, S.~Schliebs, S.~Matsuda, and N.~Kasabov, ``Span: Spike pattern
  association neuron for learning spatio-temporal spike patterns,''
  \emph{International journal of neural systems}, vol.~22, p. 1250012, 2012.

\bibitem{b22}
Y.~Meng, Y.~Jin, J.~Yin, and M.~Conforth, ``Human activity detection using
  spiking neural networks regulated by a gene regulatory network,'' \emph{The
  2010 International Joint Conference on Neural Networks (IJCNN)}, pp. 1--6,
  2010.

\bibitem{b6}
C.~Lee, P.~Panda, G.~Srinivasan, and K.~Roy, ``Training deep spiking
  convolutional neural networks with stdp-based unsupervised pre-training
  followed by supervised fine-tuning,'' \emph{Frontiers in Neuroscience},
  vol.~12, 2018.

\bibitem{b23}
S.~R. Kheradpisheh, M.~Ganjtabesh, S.~J. Thorpe, and T.~Masquelier,
  ``Stdp-based spiking deep convolutional neural networks for object
  recognition,'' \emph{Neural Networks}, vol.~99, p. 56–67, 2018.

\bibitem{b24}
J.~Berlin and M.~John, ``R-stdp based spiking neural network for human action
  recognition,'' \emph{Applied Artificial Intelligence}, vol.~34, pp. 1--18,
  2020.

\bibitem{b25}
Z.~Bing, I.~Baumann, Z.~Jiang, K.~Huang, C.~Cai, and A.~Knoll,
  ``\BIBforeignlanguage{English}{Supervised {Learning} in {SNN} via
  {Reward}-{Modulated} {Spike}-{Timing}-{Dependent} {Plasticity} for a {Target}
  {Reaching} {Vehicle}},'' \emph{\BIBforeignlanguage{English}{Frontiers in
  Neurorobotics}}, vol.~13, 2019.

\bibitem{b15}
P.~Falez, ``Improving spiking neural networks trained with spike timing
  dependent plasticity for image recognition,'' Ph.D. Thesis, Universit{\'e} de
  Lille, 2019.

\bibitem{b27}
A.~{Karpathy}, G.~{Toderici}, S.~{Shetty}, T.~{Leung}, R.~{Sukthankar}, and
  L.~{Fei-Fei}, ``Large-scale video classification with convolutional neural
  networks,'' in \emph{2014 IEEE Conference on Computer Vision and Pattern
  Recognition}, 2014, pp. 1725--1732.

\bibitem{b16}
K.~Simonyan and A.~Zisserman, ``Two-stream convolutional networks for action
  recognition in videos,'' in \emph{Proceedings of the 27th International
  Conference on Neural Information Processing Systems - Volume 1}, ser.
  NIPS'14.\hskip 1em plus 0.5em minus 0.4em\relax Cambridge, MA, USA: MIT
  Press, 2014, p. 568–576.

\bibitem{b9}
C.~Wang, ``Dual temporal scale convolutional neural network for
  micro-expression recognition,'' \emph{Frontiers in Psychology}, vol.~8, p.
  1745, 2017.

\bibitem{b7}
Y.~{Wang}, M.~{Long}, J.~{Wang}, and P.~S. {Yu}, ``Spatiotemporal pyramid
  network for video action recognition,'' in \emph{2017 IEEE Conference on
  Computer Vision and Pattern Recognition (CVPR)}, 2017, pp. 2097--2106.

\bibitem{b14}
A.~Burkitt, ``A review of the integrate-and-fire neuron model: I. homogeneous
  synaptic input,'' \emph{Biological cybernetics}, vol.~95, pp. 1--19, 2006.

\bibitem{stdprule}
G.-q. Bi and M.-m. Poo, ``Synaptic modifications in cultured hippocampal
  neurons: Dependence on spike timing, synaptic strength, and postsynaptic cell
  type,'' \emph{Journal of Neuroscience}, vol.~18, no.~24, pp.
  10\,464--10\,472, 1998.

\bibitem{b18}
G.~Farneb{\"a}ck, ``Two-frame motion estimation based on polynomial
  expansion,'' in \emph{Image Analysis}, J.~Bigun and T.~Gustavsson, Eds.\hskip
  1em plus 0.5em minus 0.4em\relax Berlin, Heidelberg: Springer Berlin
  Heidelberg, 2003, pp. 363--370.

\bibitem{b28}
J.~Canny, ``A computational approach to edge detection,'' \emph{Pattern
  Analysis and Machine Intelligence, IEEE Transactions on}, vol. PAMI-8, pp.
  679 -- 698, 12 1986.

\bibitem{b12}
H.~Wang, A.~Kläser, C.~Schmid, and C.-L. Liu, ``Dense trajectories and motion
  boundary descriptors for action recognition,'' \emph{International Journal of
  Computer Vision}, vol. 103, 2013.

\bibitem{kth1}
C.~Schuldt, I.~Laptev, and B.~Caputo, ``Recognizing human actions: A local svm
  approach,'' in \emph{Proceedings of the Pattern Recognition, 17th
  International Conference on (ICPR'04) Volume 3 - Volume 03}, ser. ICPR
  '04.\hskip 1em plus 0.5em minus 0.4em\relax USA: IEEE Computer Society, 2004,
  p. 32–36.

\bibitem{weizmann1}
L.~Gorelick, M.~Blank, E.~Shechtman, M.~Irani, and R.~Basri, ``Actions as
  space-time shapes,'' \emph{Transactions on Pattern Analysis and Machine
  Intelligence}, vol.~29, no.~12, pp. 2247--2253, December 2007.

\end{thebibliography}

\vspace{12pt}
\color{red}

\end{document}